\RequirePackage{fix-cm}
\documentclass[twocolumn]{svjour3}          
\smartqed  

\usepackage[utf8]{inputenc} 
\usepackage[T1]{fontenc}    
\usepackage{url}            
\usepackage{booktabs}       
\usepackage{amsfonts}       
\usepackage{nicefrac}       
\usepackage{microtype}      
\usepackage{xcolor}         
\usepackage{soul}
\usepackage{color, xcolor}
\soulregister{\cite}7
\usepackage{epsfig}
\usepackage{graphicx}
\usepackage{amsmath}
\usepackage{amssymb}
\usepackage{multirow}
\usepackage{color, colortbl}
\usepackage{subcaption}
\usepackage{subfloat}
\usepackage{tabularx}
\usepackage[export]{adjustbox}
\usepackage{threeparttable}
\usepackage{pifont}
\usepackage[misc]{ifsym} 
\usepackage{algorithm}
\usepackage{algpseudocode}
\usepackage{paralist}
\usepackage{listings} 
\usepackage{amssymb}
\usepackage{bbding}

\usepackage{url}
\makeatletter
\newcommand\footnoteref[1]{\protected@xdef\@thefnmark{\ref{#1}}\@footnotemark}
\makeatother

\definecolor{COLOR_CSID}{HTML}{e0f5ff}
\definecolor{COLOR_NEAROOD}{HTML}{ffefe0}
\definecolor{COLOR_FAROOD}{HTML}{ffdebf}
\definecolor{COLOR_MEAN}{HTML}{f0f0f0}

\usepackage{xspace}
\usepackage{comment}
\usepackage{bm}
\usepackage{wrapfig}

\usepackage[colorlinks,linkcolor=red,anchorcolor=blue,citecolor=blue,CJKbookmarks=True]{hyperref}

\definecolor{citecolor}{HTML}{0071BC}
\definecolor{linkcolor}{HTML}{ED1C24}

\makeatletter
\renewcommand\paragraph{
  \@startsection{paragraph} 
  {4} 
  {\z@} 
  {.5em \@plus1ex \@minus.2ex} 
  {-1.5em} 
  {\normalfont\normalsize\bfseries} 
}
\DeclareRobustCommand\onedot{\futurelet\@let@token\@onedot}
\def\@onedot{\ifx\@let@token.\else.\null\fi\xspace}

\def\etal{\emph{et al}\onedot}
\makeatother

\begin{document}

\sloppy 


\title{Learning Spatiotemporal Inconsistency via Thumbnail Layout for Face Deepfake Detection}
\author{Yuting Xu$^{1,3}$ \and 
        Jian Liang$^{2,4\star}$ \and
        Lijun Sheng$^{2,5}$ \and
        Xiao-Yu Zhang$^{1,3\star}$
}
        
\institute{
    Yuting Xu (xuyuting@iie.ac.cn) \\ 
    Jian Liang (liangjian92@gmail.com) \\
    Lijun Sheng (slj0728@mail.ustc.edu.cn) \\
    Xiao-Yu Zhang (zhangxiaoyu@iie.ac.cn) \\
    1. Institute of Information Engineering, Chinese Academy of Sciences \\
    2. CRIPAC \& MAIS, Institute of Automation, Chinese Academy of Sciences  \\
    3. School of Cyber Security, University of Chinese
Academy of Sciences \\
    4. School of Artificial Intelligence, University of Chinese Academy of Sciences \\
    5. Department of Automation, University of Science and Technology of China \\
    $\star$ Corresponding authors.\\
    }
    
\date{Received: date / Accepted: date}


\maketitle

\begin{abstract}

The deepfake threats to society and cybersecurity have provoked significant public apprehension, driving intensified efforts within the realm of deepfake video detection.
Current video-level methods are mostly based on {3D CNNs} resulting in high computational demands, although have achieved good performance.
This paper introduces an elegantly simple yet effective strategy named Thumbnail Layout (TALL), which transforms a video clip into a pre-defined layout to realize the preservation of spatial and temporal dependencies. 
This transformation process involves sequentially masking frames at the same positions within each frame.
These frames are then resized into sub-frames and reorganized into the predetermined layout, forming thumbnails. 
TALL is model-agnostic and has remarkable simplicity, necessitating only minimal code modifications.
Furthermore, we introduce a graph reasoning block (GRB) and semantic consistency (SC) loss to strengthen TALL, culminating in TALL++. 
GRB enhances interactions between different semantic regions to capture semantic-level inconsistency clues.
The semantic consistency loss imposes consistency constraints on semantic features to improve model generalization ability.
Extensive experiments on intra-dataset, cross-dataset, diffusion-generated image detection, and deepfake generation method recognition show that TALL++ achieves results surpassing or comparable to the state-of-the-art methods, demonstrating the effectiveness of our approaches for various deepfake detection problems. 
The code is available at \url{https://github.com/rainy-xu/TALL4Deepfake}.

\end{abstract}

\section{Introduction}

Generative techniques~\cite{goodfellow2014,karras2019style} have shown remarkable success across various domains. 
However, the misuse of generative algorithms for malicious purposes has given rise to a concerning phenomenon known as deepfake~\cite{kwon2021kodf,hong2022depth}. 
Deepfakes have the potential to undermine trust and security on a significant scale, posing multifaceted threats including financial fraud, identity theft, and celebrity impersonation~\cite{Verdoliva2020Media,Mirsky2021Creation}. 
With the rapid evolution of social media, the spread of deepfakes has increased, underscoring the urgent necessity for the development and implementation of robust deepfake detection techniques to effectively confront this crucial challenge.

The initial approach to deepfake video detection relied on image-level detection algorithms~\cite{a:2018,FaceXray}. 
These methods analyzed videos frame by frame and subsequently aggregated the results into a video detection output. 
However, their neglect of temporal information results in unsatisfactory performance on the increasingly sophisticated deepfake algorithms.
Therefore, recent research on deepfake detection has shifted its focus toward video-level methods.
They aim to capture spatiotemporal inconsistencies caused by frame-by-frame manipulation through spatiotemporal modeling.
Existing video-level deepfake methods generally fall into two main categories. 
One category employs dual-branch modules~\cite{masi2020two}, which separately learn spatial and temporal information before combining them.
These approaches necessitate the design of distinct fusion methods for various forgery categories and require particular attention to detail.
The other category is based on networks with temporal modeling capabilities, such as 3DCNN or Video Transformer~\cite{khormali2022dfdt}. 
Although temporal models have enhanced detection capabilities, they require significant computational resources and temporal consumption.
Thus, we wonder if there exists an efficient methodology for detecting inconsistencies in deepfake video detection.

\begin{figure}[t]
\centering
\includegraphics[width=0.97\linewidth]{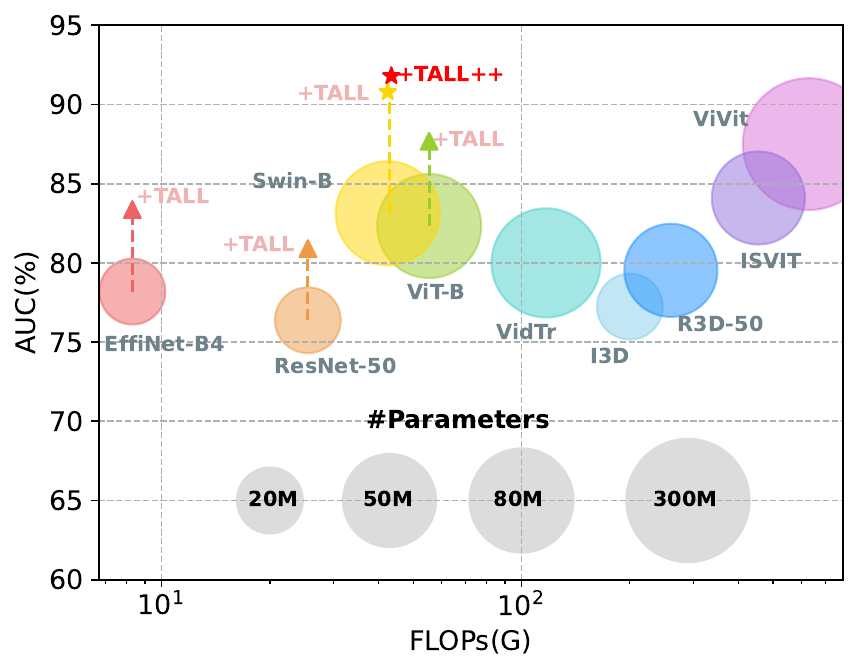}
\caption{\textbf{The AUC, FLOPs, and the number of parameters trade-off of different backbones on CDF.} 
Swin-B+TALL enjoys a better AUC-cost trade-off than 2DCNN family+TALL, 3DCNN family and most video-based visual transformers. Swin-B+TALL++ achieves state-of-the-art on CDF.
The dashed arrows depict the adaptation of TALL to each backbone, resulting in an enhancement of AUC without an increase in the number of parameters.
All models with the same setting are trained on the FF++ (HQ) dataset.
}

\label{fig:intro}
\end{figure}

In order to tackle the aforementioned challenges, we present a simple yet efficient approach for deepfake detection, which we refer to as Thumbnail Layout (TALL). 
{TALL transforms the video clips into images that retain temporal information}, converting temporal dependencies into spatial dependencies.
This innovative approach allows us to perform face deepfake video detection using image-level architecture. 
Specifically, we employ dense sampling to extract multiple clips from the video, selecting four consecutive frames at random within each clip. Subsequently, we apply a fixed-position block masking to each frame. 
Finally, we resize these frames as sub-images and arrange them sequentially into a predetermined layout resembling a thumbnail, matching the size of the clip frames.

In order to improve TALL's performance on low-quality images or new deepfake methods, we further introduce Graph Reasoning Block (GRB) and Semantic Consistency (SC) loss to form TALL++.
{GRB captures interactions between each node corresponding to specific semantic features by graph convolution, improving the interactions with different semantic regions for capturing inconsistency clues between forgery regions and real regions.}
The SC Loss calculates the similarity of temporal semantic features from consecutive frames to ensure semantic consistency and improve the generalization ability of the model.
TALL++ offers two distinct advantages over prior spatiotemporal modeling methods. 
First, TALL++ converts video tasks into image tasks, containing spatial and temporal information. This transformation enables the model to effectively capture spatiotemporal inconsistencies. 
Second, TALL++ is a model-agnostic method for spatiotemporal modeling deepfake patterns with a small increase in the amount of parameters and calculations. 
As shown in Fig.~\ref{fig:intro}, we illustrate the trade-off of AUC, Flops, and the number of parameters on the CDF dataset to evaluate the effectiveness of TALL and TALL++ combined into several common deepfake detection backbones, all achieving AUC improvement.

Experimental results on multiple benchmark datasets clearly demonstrate the proposed TALL and TALL++ obtain competitive results with the state-of-the-art, or outperform the state-of-the-art for three different cases, i.e., intra-dataset detection, cross-dataset detection, and deepfake generation method recognition problems. 
Especially, TALL++ based on swin transformer (Swin-B)~\cite{liu2021swin} achieves 91.96$\%$ AUC on cross-dataset task, FaceForensics++ $\to$ Celeb-DF~\footnote{{ The model trained on the FaceForensics++ dataset and evaluated on Celeb-DF, the same applies to the following text.}}, 78.51$\%$ on FaceForensics++ $\to$ DFDC. Our main contributions are summarized as follows.

\begin{itemize}
    \item We propose a novel strategy, Thumbnail Layout (TALL), for face deepfake video detection, which provides a new perspective for video tasks so that the model input has spatiotemporal information.
    \item TALL effectively transforms {the task of temporal modeling}  into the task of temporal modeling into the task of capturing spatial dependencies between different pixels across multiple 2D frames.
    \item We introduce graph reasoning blocks to enhance interactions with valuable semantic features while reducing redundant interactions with irrelevant features. This contributes to improving the model's resistance to perturbations and enhancing its generalization capabilities.
    \item We introduce a semantic consistency loss to enforce semantic coherence between consecutive adjacent frames, thereby enhancing the model's generalization ability. 
    \item Experiments on several datasets demonstrate our methods yield results comparable to or outperforming the state-of-the-art for three deepfake detection scenarios and even detection on diffusion model generated images.
\end{itemize}

This paper builds upon our previous work~\cite{xu2023tall} with the following enhancements:
1) Within the TALL~\cite{xu2023tall}, we additionally introduce the graph reasoning block to {enhance the relationships between the semantic regions, thereby capturing semantic-level inconsistency clues.}
2) We present the semantic consistency Loss, designed to constrain semantic relations between high-level semantic features from neighboring frames.
3) Our expanded experimental evaluation includes a broader array of datasets, covering various scenarios, including Wild-DF~\cite{zi2020add} for intra-dataset evaluation, and Deepfakes LSUN-Bedroom~\cite{ricker2022towards} for both intra-dataset and cross-dataset of diffusion-generated image detection.
4) We further evaluate our methods for the multi-classification task of deepfake method recognition on the KoDF~\cite{kwon2021kodf} dataset.
5) We conduct supplementary experiments to deepen our understanding of TALL and TALL++. We also provide in-depth modal analysis to comprehensively evaluate the proposed methods, including layout strategy design, ablation study, and visualization analysis. 

\section{Related Work}

\noindent
\subsection{Image-level Forgery Detection}\label{sec2sub1}

Typically, existing methods for forgery detection can be divided into two categories: image-level methods and video-level methods.
Image-level methods, as described in previous research~\cite{Li2018,Yang2019ExposingDF,jia2021inconsistency,fei2022learning,wu2022robust}, primarily focus on exploiting spatial artifacts present in deepfake images. These artifacts include variations between local regions~\cite{nirkin2021deepfake,Xu_visual,Yang_masked}, grid-like patterns in frequency space~\cite{pmlr-v119-frank20a,dong2022think,le2021add}, and disparities in global texture statistics~\cite{liu2020global}. 
These characteristics offer specific cues that can be used to differentiate deepfakes from genuine images.
F3Net~\cite{qian2020} and FDFL~\cite{li2021fdfl} employ a similar approach by utilizing frequency-aware features and RGB information to independently capture traces in different input spaces.
RFM~\cite{wang2021representative} and Multi-att~\cite{zhao2021multi}, on the other hand, propose attention-guided data augmentation mechanisms to assist detectors in identifying imperceptible deepfake indicators.
For forgery detection, Face X-ray \cite{FaceXray} and PCL~\cite{self_cons} provide effective techniques for outlining the boundaries of manipulated faces.
In the case of ICT~\cite{dong:2022}, an identity extraction module is utilized to identify identity inconsistencies within suspicious images.
Similarly, M2tr~\cite{wang:2022} and CORE~\cite{ni2022core} detect local inconsistencies at different spatial levels within individual frames.
In general, image-level methods are prone to overfitting and often disregard temporal information when specific image manipulation techniques are used. 
Our method preserves temporal information in thumbnails and models spatiotemporal consistency on an image-level model.

\subsection{Video-level Forgery Detection}\label{sec2sub2}

In order to enhance the effectiveness of deepfake detectors, numerous studies~\cite{Chen_2022_CVPR,shiohara2022detecting,gerstner2022detecting,zhou2021joint,amerini2019deepfake} have focused on generating diverse and generic deepfake data. Simultaneously, other studies~\cite{hu2022finfer,zheng:2021,haliassos2021lips,cozzolino2021id} have identified temporal incoherence in fake videos as a key indicator. Recent research works~\cite{haliassos2022realforensics,agarwal2019protecting} have proposed various methods to detect temporal inconsistencies using well-designed spatiotemporal neural networks. Additionally, ~\cite{gu2021spatiotemporal,gu2022delving} have attempted to incorporate temporal information by adding modules to image models.
STIL~\cite{gu2021spatiotemporal} formulates deepfake video detection as a learning process that identifies both spatial and temporal inconsistencies within a unified 2D CNN framework.
FTCN~\cite{zheng:2021} focuses on detecting temporal-related artifacts instead of spatial artifacts to enhance generalization.
LipForensics~\cite{haliassos2021lips} aims to learn high-level semantic irregularities in mouth movements within generated videos.
RealForensics~\cite{haliassos2022realforensics} utilizes auxiliary data sets during training to improve generalization, albeit at the expense of increased computational demands.
While video-based methods achieve strong generalization, they often come with significant computational overhead.
To mitigate computational costs, we propose TALL, which involves aggregating consecutive video frames into thumbnails for learning spatiotemporal consistency.

\subsection{Deepfake Detection with Visual Transformer}\label{sec2sub3}

Lately, the Vision Transformer (ViT)~\cite{vit} has demonstrated remarkable performance in computer vision tasks. Several studies have extended ViT for forgery detection~\cite{wodajo:2021,heo:2021,10.1145/3512732.3533582,coccomini:2022,wodajo:2021,heo:2021}, surpassing CNN-based models~\cite{sun2022dual,yu2020responsible,yu2022improving} in terms of performance. 
However, these methods often sacrifice computational efficiency.
M2tr~\cite{wang:2022} introduces a multi-modal multi-scale transformer that detects local inconsistencies at different scales and leverages frequency features to improve the robustness.
Zhao et.al.~\cite{zhao:2022}, propose a self-supervised transformer-based audio-visual contrastive learning method, which learns mouth motion representations by encouraging the paired video and audio representations to be close while unpaired ones to be diverse.
While ICT~\cite{dong:2022} focuses on detecting identity consistency in deepfake videos, it may struggle to detect face reenactment and entirely synthesized faces. DFLL~\cite{khan:2021} utilizes the UV texture map to aid the transformer in detecting deepfakes, but this approach may disrupt the continuity between video frames. DFTD~\cite{khormali2022dfdt} employs ViT to consider both global and local information, yet it overlooks the issue of excessive model arithmetic requirements.

Although transformer-based approaches~\cite{wodajo:2021,sun2022faketransformer} exhibit promising performance, their computational complexity poses challenges in deployment and usage. Additionally, long-range dependencies may not be adequately exploited in these detection models.
To address these challenges, we collaborate with the Swin Transformer~\cite{liu2021swin}, which produces a hierarchical feature representation with linear computational complexity relative to the input image size. This property makes it a suitable backbone for various vision tasks. 
Differing from two-branch architectures~\cite{r:2021,feichtenhofer2019slowfast}, our approach captures both short-range and long-range temporal inconsistencies using a single-branch model. 
This is made possible by leveraging the transformative power of the Swin Transformer to model long-range data. 

\begin{figure*}[ht]
\centering
\includegraphics[width=0.99\linewidth]{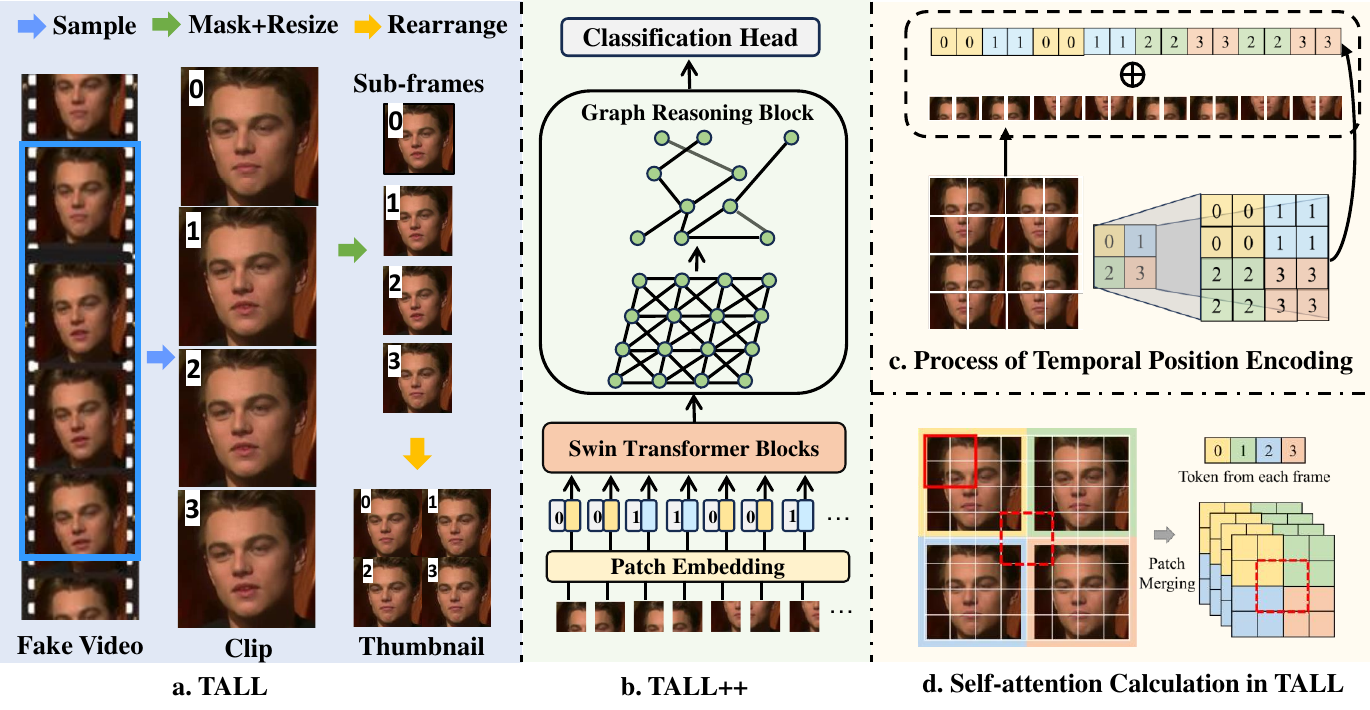}
\caption{(a) TALL formation process. 
For the sake of simplicity, we omit the masking procedure in this representation. (b) The pipeline of TALL++. 
Here we use the Swin transformer as the backbone to illustrate the subsequent process. 
First, the thumbnail images are flattened and subsequently supplemented with temporal position encoding before being input into the Swin Transformer Blocks (STB). 
Following this, {the Graph reasoning block takes the output of STB to enhance relations of valuable features}. Ultimately, the classification head generates predictions.
(c) Illustration of the formation of temporal position encoding of TALL++. (d) Illustration of the calculation of self-attention in TALL.}
\label{tall_pipe}
\end{figure*}

\subsection{Graph Reasoning}\label{sec2sub4}
Graph reasoning (GR) has gained popularity and has proven to be a practical approach to reasoning about relationships. 
GR starts with projecting high-level semantic features in coordinate space into nodes in interaction space. 
Then perform relation reasoning over the graph via a graph convolution network (GCN)~\cite{kipf2016semi}. 
Upon analyzing the graph, the node features are projected back into the coordinate space. 
Wang $et.al$~\cite{wang2018videos} propose a GCN model for reasoning with multiple relationships between different objects in a long-range video.  
Following the work, GR involves a structured dense connectivity graphs~\cite{wang2018non,liang2018symbolic,chen2019graph} that were proposed and successfully applied in a variety of computer vision applications, such as visual reasoning~\cite{yang2022psg} and forgery detection~\cite{Cao_2022_CVPR}. 
GR has demonstrated immense potential in effectively capturing and modeling long-range dependencies, which play a crucial role in numerous computer vision tasks. According to our task, GR can enhance interactions of inter-class features and reduce redundant interactions of intra-class features.

\subsection{Videos as a stack of still images}
{
Many efforts have been made to tackle video tasks through an image-based perspective.
Early works like~\cite{davis1997representation,yao2012action} rely on static images to infer target actions. 
Davis~\etal~\cite{davis1997representation} identifies multiple small objects in the image and infers from their relationships.
Bilen~\etal~\cite{bilen2016dynamic} attempt to summarize RGB or motion information in a video into a representative image called dynamic image.
STCNN~\cite{safaei2019still} predicts missing temporal information in static images and combines it with spatial information for action classification.
In contrast, our method does not strive to comprehend a video via a solitary input image or a compact image. 
Rather, we propose a method that transforms the video into a comprehensive thumbnail and effectively classifies the image for video deepfake detection.}

\section{Methodology}

Our objective is to tackle the problem of detecting face deepfake in videos using the TALL strategy, {which captures temporal inconsistencies by converting video clips into thumbnails containing temporal and spatial information.}
First, we use the dense sampling strategy to obtain several video clips. 
Second, we convert consecutive frames into thumbnail images in order to transform the temporal modeling between frames into spatial modeling, where the model captures the dependency of the exact pixel positions between different frames.
Subsequently, the backbone model incorporates these thumbnail images to identify spatiotemporal inconsistencies across frames.
Then we introduce a graph reasoning block for TALL {to enhance relationships between the semantic regions, enabling the model to capture inconsistent cues at the semantic level.}
Finally, we implement a semantic consistency (SC) loss to constrain the consistency of adjacent frames. 
Applying such a GRB and SC loss to TALL yields TALL++. 
In the following, we provide a detailed explanation of each step.

\subsection{Thumbnail Layout}

While slight motion blurring has been employed in recent studies to tackle the apparent shortcomings of the deepfake technique, subtle spatio-temporal artifacts persist. 
These artifacts play a crucial role in detecting deepfakes. However, there are a few issues that arise when using models with temporal modeling capabilities. 
First, the computation cost is large.
In addition, it is important to note that when analyzing videos over a long period of time, there may be small artifacts or nuances that are overlooked or not taken into account. 
In order to overcome these challenges, we propose a TALL strategy, which incorporates temporal information into image-level tasks while preserving spatial information intact.
TALL is a model-agnostic method. In order to better explain the mechanism of TALL, we mainly use Swin-B~\cite{liu2021swin} as the backbone to explain TALL in this section and subsequent sections.
As shown in Fig~\ref{tall_pipe}(a), {given a video $V\in\mathbb{R}^{{T}\times{C}\times{H}\times{W}}$, where $T$ is the number of frames ($T=4$ by default), $C$ is the number of channels, and $H\times{W}$ is the size of frames.  
Then we utilize a masking strategy on each frame and resize each frame to obtain sub-frames of size ${C}\times{\frac{H}{\sqrt{T}}}\times{\frac{W}{\sqrt{T}}}$. 
Subsequently, $T$ sub-frames are rearranged as Thumbnail $I\in\mathbb{R}^{{C}\times{H}\times{W}}$.
In this process, we utilize four consecutive frames to illustrate the rearrangement. 
These frames are rearranged into a 2$\times$2 thumbnail based on the coordinates [0,0], [0,1], [1,0], [1,1].}

{
The masking strategy is a data augmentation designed specifically for thumbnails, improved from Cutout~\cite{devries2017improved}. 
The specific approach is as follows: take $T$ frames as a unit, select a fixed-size square area at the same position within each frame of the unit, and fill it with zeros. 
The erased position of the next unit will change.
There are two main reasons for this strategy: first, having different mask positions between thumbnails promotes the network to pay more attention to complementary and less prominent features. 
Second, having the same mask area within the sub-frames of the thumbnail allows the model to better focus on subtle changes between adjacent frames, thus capturing the inconsistencies introduced by the frame-wise manipulation in deepfake videos. The detailed procedures of TALL and masking strategy are shown in Algorithm~\ref{alg_tall}.}

\begin{algorithm}[ht]
\caption{Pseudocode of TALL}
\label{alg_tall}
\algnewcommand{\LeftComment}[1]{\Statex \(\triangleright\) #1}
\noindent\textbf{Input:} $T$ video frames $F_t\in\mathbb{R}^{{C}\times{H}\times{W}}$, \\
Size of mask: $S\times{S}$ 

\noindent\textbf{Output:} thumbnail image $I\in\mathbb{R}^{{C}\times{H}\times{W}}$

\begin{algorithmic}[1]
\State Randomly sample x,y from W,H  
\State $M$ = Initialize$\_${matrix}$\_${with}$\_${all}$\_${ones}$(H, W)$ 
\State $M[x:x+S, y:y+S] = 0$
\State Initialize$\_$masked$\_$frames$\_$List: L
\For{$t=1, \cdots, T$}
    \For{$i=1,\cdots,C$}
        \State $F_m = F_t \cdot M$
        \State $f_{down} = $Downsampling$(f_m)$
        \State Append $f_{down}$ to $L$
    \EndFor
\EndFor
\LeftComment {Rearrange like Fig. 1(a).}
\State $I$ = Rearrange$(L,`(TC)\times{H}\times{W}\to C\times{\sqrt{T}}{H}\times{\sqrt{T}{W}}')$
\end{algorithmic}
\end{algorithm}

\subsection{Scaling TALL to Vision Transformer}

\begin{figure}[ht]
\centering
\includegraphics[width=1\linewidth]{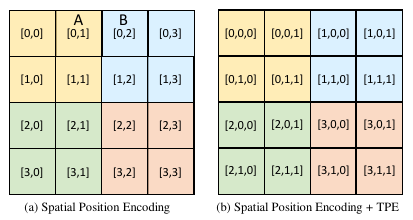}
\caption{Illustration of the difference between spatial position encoding and spatial position encoding+TPE. }
\label{fig:pe}
\end{figure}

Given a thumbnail image {$I\in\mathbb{R}^{{C}\times{H}\times{W}}$}, we reshape the images $I$ into a sequence of flattened image patches {$I_p\in\mathbb{R}^{N\times{P^{2}{c}}}$, where (P, P) is the size of each patch, and $N = HW/P^{2}$
is the resulting number of patches, and $c$ is the embedding dimension. }
{Due to the original spatial position encoding can mislead the model to think that pixels A and B are adjacent, but in reality, A and B are from different image frames, as shown in Fig.~\ref{fig:pe}(a).
Accordingly, we enhance the original spatial position encoding by adding our Temporal Position Encoding (TPE).}
As shown in Fig.~\ref{tall_pipe}(c), to begin, we randomly initialize four position encodings representing patches from different frames to contain temporal information. 
Subsequently, we broadcast these four position encodings to match the size of image patches $I_p$. The $I_p$ is then summed with the TPE and fed into Swin transformer blocks (STB). 
{By adding TPE to spatial encoding, positional encoding adds a dimension of information to represent the temporal order of frames, as shown in Fig.~\ref{fig:pe}(b).
Note that Swin Transformer~\cite{liu2021swin} incorporates spatial positional encoding into the attention matrix.
}

Thereafter, the high-level semantic characteristics yielded by STB are introduced to GRB to acquire more illustrative attributes and eliminate superfluous features.

Here we use the shifted window to explain the framework's mechanism. 
As illustrated in Fig.~\ref{tall_pipe} (d), the model calculates self-attention and takes into account spatial relationships between sub-frames (shown as the solid red box). 
If the window covers multiple sub-frames (shown as the red dashed box), the model can detect differences over time between frames. TALL uses both local and global contexts of deepfake patterns to accurately model short and long-range spatial dependencies. Compared to prior approaches, TALL++ balances speed and accuracy by sacrificing some spatial information while maintaining a promising performance.

\begin{figure}[ht]
\centering
\includegraphics[width=0.95\linewidth]{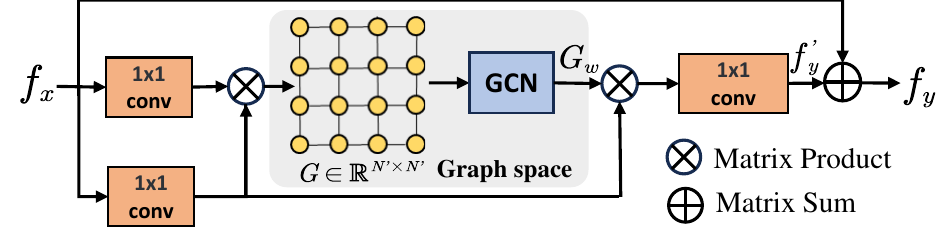}
\caption{The structure of graph reasoning block and the procedure of obtaining weighted feature $f_y$.}
\label{fig:grb}
\end{figure}

Besides, we make some adjustments to the window size of Swin-B based on TALL++'s characteristics. 
We increased the window size of the first three stages to encourage more frequent interaction between frames in the thumbnail, allowing the model to learn more detailed spatio-temporal dependencies. 
We also set the window size of the final stage to match the feature map size, which enables global attention computations while TALL++ captures global spatial-temporal dependencies. 
As a result, the size of the last layer of the feature map decreased, reducing the window size without adding any extra computational load. The window sizes for the four stages of Swin-B are $[14,14,14,7]$.

\subsection{Graph Reasoning Block}

{
Given the universal video perturbations on social media, detection methods that only capture pixel inconsistencies would be significantly impacted. Therefore, we utilize a Graph Reasoning Block (GRB) to enhance the relationships between visual semantic regions (background, hair, face, etc.), thereby capturing semantic-level inconsistency clues.}

{
As shown in Fig.~\ref{fig:grb}, GRB takes the embedded features $f_x\in\mathbb{R}^{{N'}\times{d}}$ as the input, which are the features calculated by the Swin Transformer Block before average pooling. 
$N'$ is the number of features, $d$ is the dimension of a feature.
These features can be seen as highly concentrated semantic regions.
GRB treats these features as nodes in a fully connected graph, views the interactions between features as edges, and determines the weights of these interactions through learning.
We first map original features $f_x$ from the coordinate space into the adjacent matrix $G\in\mathbb{R}^{{N'}\times{N'}}$ in graph space though $1\times{1}$ convolution layer.}

{After feature projection, the model calculates interactions between each node by graph convolution network(GCN)~\cite{kipf2016semi}. 
GCN updates and calculates the weights of the edges between nodes in the adjacency matrix. Eventually, we obtain a weighted adjacency matrix $G_w$ that contains information about the strength of relationships.
For example, if two nodes represent semantic information from the eye and mouth regions in the forged image respectively, and the mouth region has been counterfeited, then the weights of the edges between these two nodes would increase.
Finally, we re-project the weighted adjacency matrix $G_w$  back into the coordinate space, obtaining the features $f'_y$ with relationship information of high-level semantic regions. By adding $f'_y$ to the original feature $f_x$, we get the final output $f_y$ of the module.
}

\subsection{Semantic Consistency Loss}
As aforementioned, high-level semantic features help the model improve generalization and robustness. 
Specifically, when images are compressed or perturbed, pixel information is destroyed, while high-level semantic information is less affected.
Hence, we propose a SC loss to penalize the distance between the high-level semantic features produced by GRB. We adopt MSE Loss (${L}_{mse}$) to penalize the distance between features from two adjacent frames:
\begin{equation}
\begin{aligned}{
\mathcal{L}^t_{mse}=\frac{1}{N}\sum_{i=1}^{N}(f^{t}_{i}-f^{t-1}_i)^2,
}
\end{aligned}
\end{equation}
where $N$ denotes the feature size, $t$ denotes the $t$-th frame. For $T$ successive frames, the multi-temporal consistency loss can be written as:

\begin{equation}
\begin{aligned}{
\mathcal{L}_{sc} =\frac{1}{T-1}\sum_{t=2}^{T}\mathcal{L}^t_{mse}.
}
\end{aligned}
\end{equation}

\noindent\textbf{Classification loss.} 
We use the standard cross-entropy loss as classification loss:
\begin{equation}
\small
    \begin{aligned}{
        \mathcal{{L}}_{ce} = - \frac{1}{n} \sum_{i=1}^{n}{y_i \log{\mathcal{F}(x_i)} 
        + (1-y_i) \log{(1-  \mathcal{F}(x_i)})},
    }
    \end{aligned}
\end{equation}
where $x_i$ indicates input clip,  $y_i$ denotes the label of clip, $n$ is the number of clip, $\mathcal{F}$ is TALL++.

\noindent\textbf{Overall loss.} For the objective function of our framework, we combine the SC loss with the CE loss to form the overall loss in our framework:

\begin{equation}
\small
    \begin{aligned}{
        \mathcal{L} =\mathcal{L}_{ce}+ \alpha \mathcal{L}_{sc}.
    }
    \end{aligned}
\end{equation}
By default, we set $\alpha= 0.5$ in our experiments.

\section{Experiment}

To assess the effectiveness of our proposed approach, we conduct a comprehensive set of experiments. 
Our evaluation contains three distinct deepfake detection scenarios: intra-dataset detection, cross-dataset detection, and detection under various disturbances, across multiple well-established benchmarks. 
Furthermore, we extend the evaluation to include TALL++ on a widely used dataset of diffusion-generated images and introduce a new task focused on deepfake method recognition using the KoDF dataset. 
Additionally, we analyze the efficacy of each constituent element within TALL++ through ablation studies and explore its performance under various parameter configurations.

\subsection{Setup}
\textbf{Datasets.} Following previous works~\cite{haliassos2021lips,haliassos2022realforensics,zheng:2021}, we evaluate our method on seven widely used datasets.

\textbf{FaceForensics++ (FF++)}~\cite{FaceForensics} is a most-used benchmark on intra-dataset deepfake detection, consisting of 1,000 real videos and 4,000 fake videos in four different manipulations: DeepFake{ 
 (DF)}~\cite{deepfake-faceswap}, FaceSwap{ 
 (FS)}~\cite{faceswap}, Face2Face{ 
 (F2F)}~\cite{thies2016face2face}, and NeuralTextures{ (NT)}~\cite{Thies:2019}. Besides, FF++ contains three video qualities, high quality (HQ), low quality (LQ), and RAW. 

\textbf{Celeb-DF (CDF)}~\cite{li2020celeb} is a popular benchmark on cross-dataset, which contains 5,693 deepfake videos generated from celebrities. The improved compositing process was used to improve the various visual artifacts presented in the video. 

\textbf{DeepFake Detection Challenge (DFDC)}~\cite{dfdc} is a large-scale benchmark developed for Deepfake Detection Challenge. This dataset includes 124k videos from 3,426 paid actors. The existing deepfake detection methods do not perform very well on DFDC due to their sophisticated deepfake techniques. 

{\textbf{FaceShifter (FSh)}~\cite{li2020advancing} generates the swapped face with high-fidelity by exploiting and integrating the target attributes thoroughly and adaptively based on 1,000 original videos from FF++. }

\textbf{DeeperForensics (DFo)}~\cite{jiang2020deeperforensics} includes 60,000 videos with 17.6 million frames for deepfake detection, whose videos vary in identity, pose, expression, emotion, and lighting conditions, etc. Besides, apply random-type distortions to the 1, 000 raw manipulated videos at five different intensity levels, producing a total of 5, 000 manipulated videos.

\textbf{Wild-Deepfake (Wild-DF)}~\cite{zi2020add} encompasses 7,314 facial sequences extracted from a total of 707 deepfake videos gathered from online sources. Wild-DF presents a wealth of diversity in scenes and features a higher presence of individuals within each context.

\textbf{KoDF}~\cite{kwon2021kodf} is a large-scale deepfake detection dataset, containing 175,776 fake clips and 62,166 real clips of 403 Korean subjects.  The dataset employs six different models to generate deepfake clips: FaceSwap, DeepFaceLab, FSGAN, First Order Motion Model (FOMM), Audio-driven Talking Face Head Pose (AD), and Wav2Lip. 

\textbf{Deepfakes LSUN-Bedroom (DLB)}~\cite{ricker2022towards} is a comprehensive diffusion-generated and GAN-generated benchmark including images generated by five diffusion models (DDPM, IDDPM, ADM, PNDM and LDM) and five GANs (ProGAN, StyleGAN, ProjectdedGAN, Diff-StyleGAN2 and Diff-ProjectedGAN). DLB contains 50k samples in total from which 39k are used for training, 1k for validation, and 10k for testing.

\begin{table}[ht]
\centering

\caption{\textbf{Intra-dataset evaluations on FF++.} Intra-testing in terms of video-level Acc. ($\%$) and AUC ($\%$) on two quality settings, HQ indicates high quality, and LQ indicates low quality. Swin-B is used as the backbone network for TALL and TALL++. } 
\begin{tabular}{lcccc}
\toprule
\multicolumn{1}{l}{\multirow{2}{*}{Methods\quad}} & \multicolumn{2}{c}{ FF++ (HQ)} & \multicolumn{2}{c}{FF++ (LQ)} \\ \cmidrule(l){2-5}
\multicolumn{1}{c}{} &  Acc.    &AUC    &  Acc.      &  AUC    \\ \midrule

MesoNet~\cite{a:2018}  & 83.10 & -      & 70.47   & -      \\
Xception~\cite{xception} & 95.73 & 96.30      & 86.86 & 89.30        \\
Face X-ray~\cite{FaceXray}  & -      & 87.35 & -        & 61.60 \\
Two-branch~\cite{masi2020two}  & 96.43 & 98.70 & 86.34   & 86.59 \\
Add-Net~\cite{zi2020add}  & 96.78 & 97.74 & 87.50   & 91.01 \\
F3Net~\cite{qian2020}                & 97.52 & 98.10 & 90.43   & 90.43 \\
FDFL~\cite{li2021fdfl}                  & 96.69 & 99.30 & 89.00   & 92.40 \\
Multi-Att~\cite{zhao2021multi}  & 97.60 & 99.29 & 88.69   & 90.40 \\
RECCE~\cite{Cao_2022_CVPR} &97.06 &99.32 &91.03 &95.02 \\ 
LipForensics~\cite{haliassos2021lips} &98.80 &99.70 & 94.20 & {98.10} \\  \midrule
DFDT~\cite{khormali2022dfdt}               &98.18   & 99.26      & 92.67  & 94.48     \\
ADT~\cite{wang2022adt}               & 92.05   & 96.30       & 81.48  & 82.52      \\
ST-M2TR~\cite{wang:2022}     & -   & 99.42    & -  & 95.31     \\
VTN~\cite{neimark2021video}  & 98.47 & -  &  94.02  & - \\
VidTR~\cite{zhang2021vidtr} &  97.42 & - &  92.12 & -  \\
ViViT$^*$~\cite{vivit} &   92.60  & - & 88.02 & -  \\
ISTVT~\cite{zhao2023istvt} & {\textbf{99.00}}  & -      &  96.15  & - \\
\textbf{TALL}        &  98.65 &   99.87    &   92.82   & 94.57    \\ 
\textbf{TALL++}   & 98.71  &  \textbf{99.92}   & \textbf{96.32}    &\textbf{98.68}    \\ 
\bottomrule
\end{tabular}
\label{table:ffpp}
\end{table}

\subsection{Implementation Details}
\textbf{Network hyper-parameters.} We use MTCNN~\cite{zhang2016joint} to detect face for each frame in the deepfake videos, only extract the maximum area bounding box and add 30$\%$ face crop size from each side as in LipForensics~\cite{haliassos2021lips}. The ImageNet-21K pretrained Swin-B model is used as our backbone. Excluding ablation experiments, we sample 8 clips using dense sampling, each clip contains 4 frames. The size of the thumbnail is $224\times$224. 
Adam~\cite{kingma2014adam} optimization is used with a learning rate of 1.5e-5 and batch size of 4, using a cosine decay learning rate scheduler and 10 epochs of linear warm-up. 

\textbf{Evaluation metrics.} We adopt Acc. (accuracy) and AUC (Area Under Receiver Operating Characteristic Curve) as the evaluation metrics for extensive experiments. To ensure a fair comparison, we calculate video-level predictions for the image-based method and average the predictions across the entire video.

\subsection{Intra-dataset Evaluation}
\textbf{FF++}. We first conduct experiments on the FF++ dataset under both Low Quality (LQ) and High Quality (HQ) settings, as shown in Table~\ref{table:ffpp}. 
Not surprisingly, transformer-based methods obtain better performance compared to most CNN-based methods. 
TALL obtains comparable accuracy for each task and also outperforms prior work in terms of the HQ AUC.
However, TALL falls short of achieving a desirable AUC in LQ. LipForensics~\cite{haliassos2021lips} managed to secure state-of-the-art outcomes, which may stem from its ability to capture complex high-level semantic features of lip-reading.

Compared with the other methods, TALL++ always achieves promising results.
{TALL++ beats state-of-the-art method ISVIT~\cite{zhao2023istvt} by an increment of 0.17$\%$ in accuracy under LQ settings - a notably more challenging task.}
Besides, TALL++ performs better than TALL due to the contribution of graph reasoning at the semantic level and the constraint of semantic consistency loss.
Notably, TALL++ improves the HQ accuracy from 98.65$\%$ to 98.71$\%$ and AUC from 99.87$\%$ to 99.92$\%$, the LQ accuracy from 92.82$\%$ to 96.32$\%$ and AUC from 94.57$\%$ to 98.68$\%$.
Especially in the challenging LQ setting, compared with LipForensics~\cite{haliassos2021lips}, the accuracy of TALL++ exceeds it by {2.12$\%$}. 

\begin{table}[ht]
\centering
\caption{\textbf{Intra-dataset evaluations.} We report the video-level AUC ($\%$) on CDF, DFDC and Wild-DF. } 
\begin{tabular}{lcccc}
\toprule
Methods     & CDF & DFDC & Wild-DF & Avg. \\ \midrule
Xception~\cite{FaceForensics}      &  99.44   & 84.58     & 83.25 & 89.09 \\
TEI~\cite{liu2020teinet}         & 99.12    & 86.97     & 81.64 & 89.24\\
ADDNet-3d~\cite{zi2020add}   & 95.16    &  79.66    & 65.50 & 80.11\\
RFM~\cite{wang2021representative}        & 97.96    & 80.83     & 77.38 & 85.39\\
F3Net~\cite{qian2020}      & 97.52   &  76.71    & 80.66 & 84.96\\
Multi-Att~\cite{zhao2021multi}    &97.92   &   76.81       &82.86 & 85.86\\
RECCE~\cite{Cao_2022_CVPR}    &  98.59    &  81.20     &  83.25 & 87.68\\
STIL~\cite{gu2021spatiotemporal}        & 99.78    &  89.80     & 84.12 & 91.23 \\ 
\textbf{TALL}   & 99.24    &  89.61   & 88.75  & 92.53  \\
\textbf{TALL++} & \textbf{99.98}    &  \textbf{90.68}   &\textbf{91.12}    &\textbf{93.93} \\
\bottomrule
\end{tabular}
\label{intra:cdf}
\end{table}

\textbf{CDF $\&$ DFDC $\&$ Wild-DF}. We also perform intra-dataset evaluations on the CDF, DFDC, and Wild-DF datasets. The AUC values for our methods and prior research are presented in Table~\ref{intra:cdf}.
It is worth noting that while all existing methods demonstrate satisfactory results when applied to the CDF dataset, their performance noticeably deteriorates when dealing with the DFDC and Wild-DF datasets. For instance, the state-of-the-art method, STIL~\cite{gu2021spatiotemporal}, achieves an AUC of 99.78$\%$ on CDF but experiences a decline to 89.80$\%$ on DFDC and 84.12$\%$ on Wild-DF.
As expected, TALL++ achieves the best intra-dataset AUC value on each dataset, even when ignoring the GRB and SC loss, TALL can still obtain promising AUC 90.68$\%$ on DFDC and 88.75$\%$ on Wild-DF.

\begin{figure}[htbp]
\centering
\includegraphics[width=0.45\textwidth]{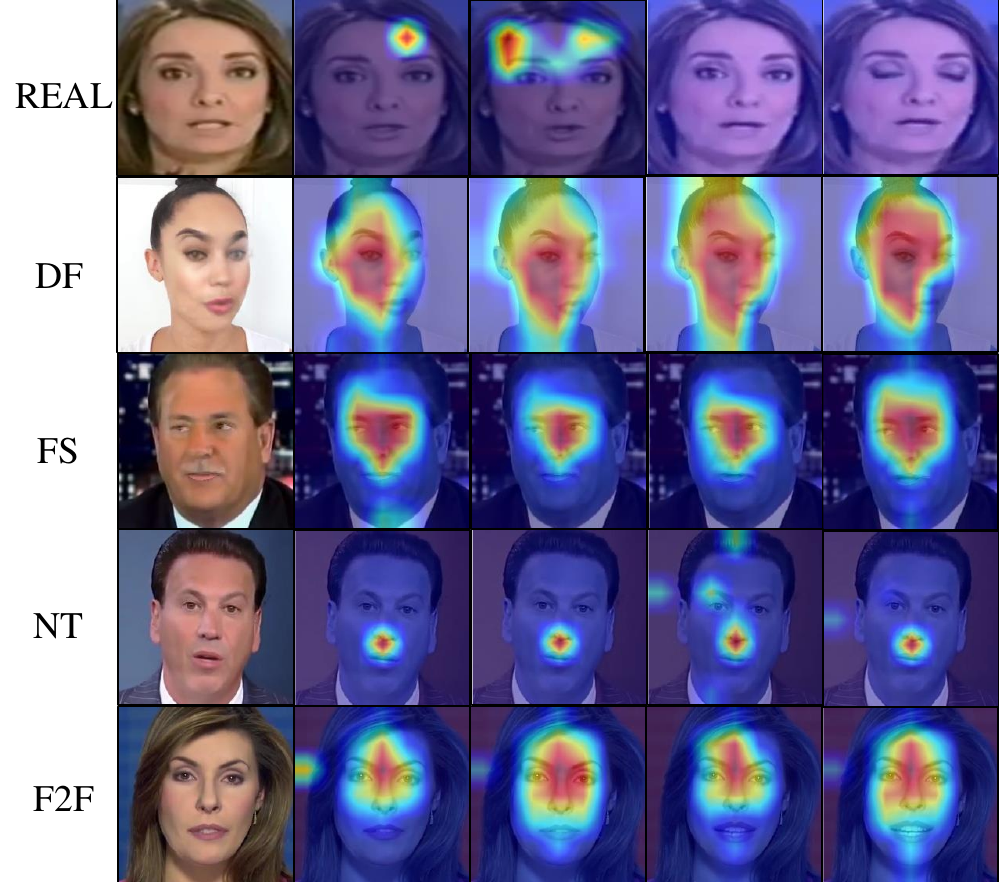}
\caption{\textbf{Saliency map visualization on FF++.} {We give the class activation maps of a clip from the real videos and four types of deepfake videos (DF, F2F, FS, NT). It can be observed that TALL++ can locate the forged areas corresponding to different forgery types.}}
\label{cam:ffpp}
\end{figure}

\textbf{Saliency map visualization on FF++}. 
We adopt Grad-CAM~\cite{Selvaraju_2017_ICCV} to visualize where the TALL++ is paying its attention to the deepfake faces. In Fig.~\ref{cam:ffpp}, we give the results on intra-dataset scenarios on FF++ (HQ). 
It should be noted that in the case of the DF, the facial region is transferred from a source video to a target, and TALL++ is specifically attuned to focus on this facial region.
In contrast, the NT exclusively alters facial expressions within the mouth region, and TALL++ similarly focuses on this specific mouth region.

\subsection{Cross-dataset Evaluation}

\textbf{CDF $\&$ DFDC $\&$ Fsh $\&$ DFo.} To analyze the generalization ability of our method, we test the model trained on FF++ (HQ) on several unseen datasets. As shown in Table~\ref{table:cross}, TALL performs well for three challenging tasks (FF++$\to$CDF, FF++$\to$DFDC, and FF++$\to$DFo), and obtains an average AUC 91.71$\%$ that is competitive to two state-of-the-art methods, RealForensics~\cite{haliassos2022realforensics} and ISVIT~\cite{zhao2023istvt}. Similar to the observations in intra-dataset tasks, GRB and SC Loss are beneficial to cross-dataset tasks, and TALL++ achieves state-of-the-art performance. On the medium-sized CDF dataset, our method TALL++ significantly outperforms previously published state-of-the-art approaches, advancing the AUC from 88.30$\%$ in DFDT~\cite{khormali2022dfdt} to 91.96$\%$. Besides, TALL++ gets the best result 78.51$\%$ on the large-scale challenging DFDC dataset. Generally, the TALL strategy works well enough, as seen from the outperforming results of TALL over prior methods, and the GRB and SL loss further lifts the average performance by nearly 0.6$\%$.

\begin{table}[ht]
\centering
\setlength\tabcolsep{3pt} 
\caption{ \textbf{Generalization to unseen datasets.} We report the video-level AUC ($\%$) on four unseen datasets: CDF, DFDC, FSh, and DFo. { “Avg'' indicates the mean AUC across all datasets.}}
\begin{tabular}{lcccccccccccccc}
\toprule
Method         & CDF   & DFDC  & FSh   & DFo   & Avg.      \\ 
\midrule
Xception~\cite{xception}       & 73.70 & 70.90 & 72.00 & 84.50 & 75.28    \\
CNN-aug~\cite{wang2020cnnaug}        & 75.60 & 72.10 & 65.70 & 74.40 & 71.95    \\
CNN-GRU~\cite{sabir2019gru}        & 69.80 & 68.90 & 80.80 & 74.10 & 73.40    \\
Patch-based~\cite{chai2020patch}    & 69.60 & 65.60 & 57.80 & 81.80 & 68.70    \\
Face X-Ray~\cite{FaceXray}     & 79.50 & 65.50 & 92.80 & 86.80 & 81.15    \\
Multi-Att~\cite{zhao2021multi}     & 75.70 & 68.10 & 66.00 & 77.70 & 71.88    \\
DSP-FWA~\cite{li2019DSP}        & 69.50 & 67.30 & 65.50 & 50.20 & 63.13    \\
LipF~\cite{haliassos2021lips}  & 82.40 & 73.50 & 97.10 & 97.60 & 87.65    \\
\midrule
FTCN~\cite{zheng:2021}          & 86.90 & 74.00 & 98.80 & 98.80 & 89.63 \\
RealForensics~\cite{haliassos2022realforensics} & 86.90 & 75.90 & 99.70 & 99.30 & 90.45   \\ 
DFDT~\cite{khormali2022dfdt}            & 88.30 & 76.10 & 97.80 & 96.90 & 89.70    \\
VTN~\cite{neimark2021video} & 83.20 & 73.50 & 98.70 & 97.70 & 88.30 \\
VidTR~\cite{zhang2021vidtr}  & 83.50 & 73.30 & 98.00 & 97.90 & 88.10 \\
ViViT$^*$~\cite{vivit} &86.96   & 74.61 & 99.41 &99.19 & 90.05 \\
ISTVT~\cite{zhao2023istvt}  & 84.10 & 74.20 & 99.30 & 98.60 & 89.10 \\
\textbf{TALL}          & 90.79     & 76.78   & 99.67      & 99.62    & 91.71         \\
\textbf{TALL++}          &\textbf{91.96}       &\textbf{ 78.51 }   & \textbf{99.98}      & \textbf{99.94 }   &\textbf{92.35}         \\
\bottomrule
\end{tabular}
\label{table:cross}
\end{table}

\begin{figure}[ht]
\centering
\includegraphics[width=0.43\textwidth]{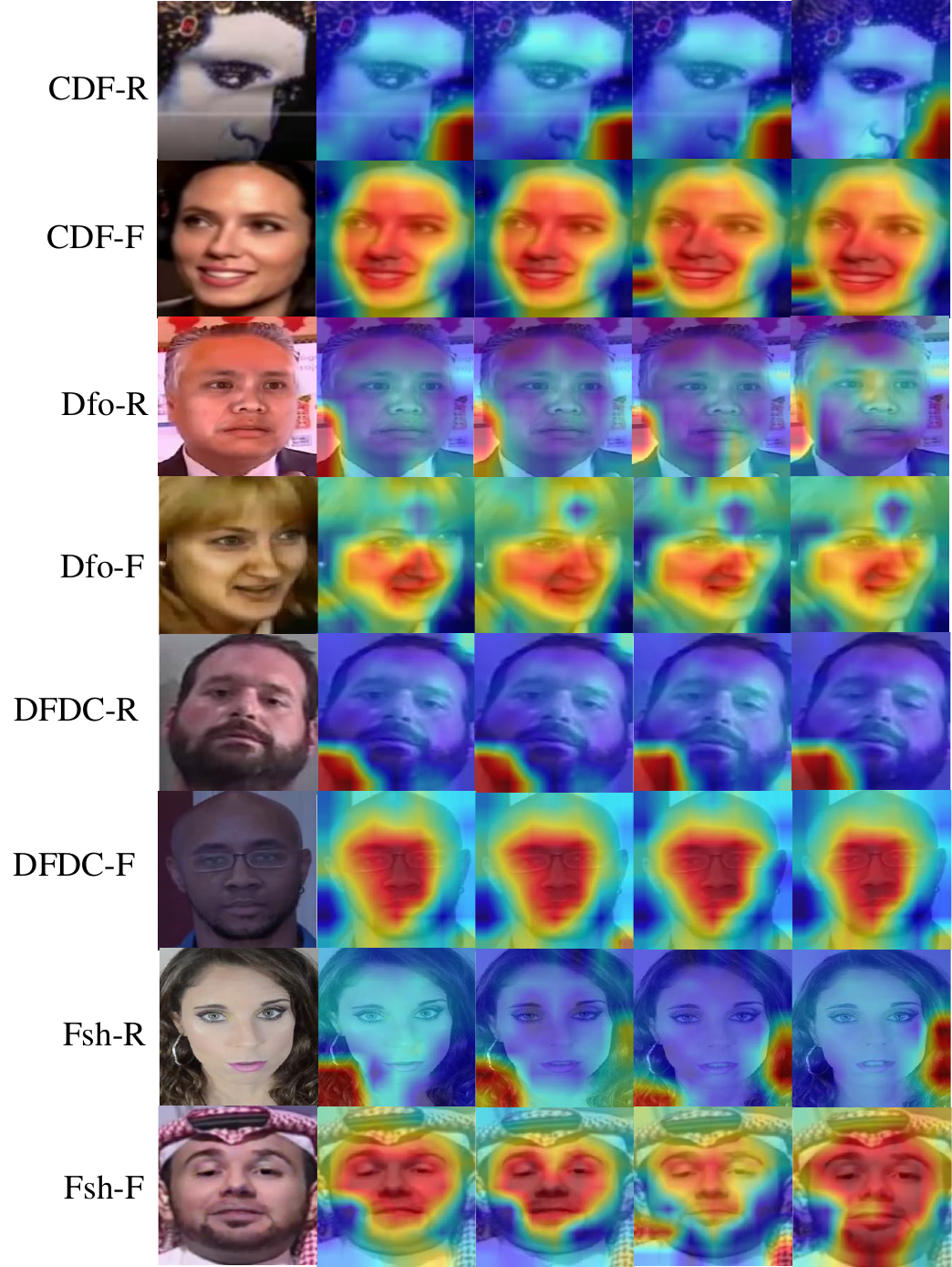}
\caption{Saliency map visualization for cross-dataset scenario on CDF, DFo, DFDC and Fsh. {We present the class activation maps of a clip from real and fake videos on each dataset. '-R' indicates a real image, while '-F' indicates a fake image.}}
\label{cam:cross}
\end{figure}

\textbf{Saliency map visualization.} 
In Fig.~\ref{cam:cross}, we visualize the saliency map for the cross-dataset scenario by training the model on FF++ (HQ). 
The figure reveals that our method generates distinct heatmaps for genuine and manipulated facial images, showing significant variations in prominent regions across different deepfake techniques, such as blending boundaries (CDF), and unusual motion patterns within the clips (DFDC, Fsh, Dfo). 
This is achieved despite our reliance on binary labels exclusively during the training phase. 
The visualization further verifies the outstanding generalization ability of our proposed methods.

\begin{figure*}[ht]
\centering
\includegraphics[width=1\textwidth]{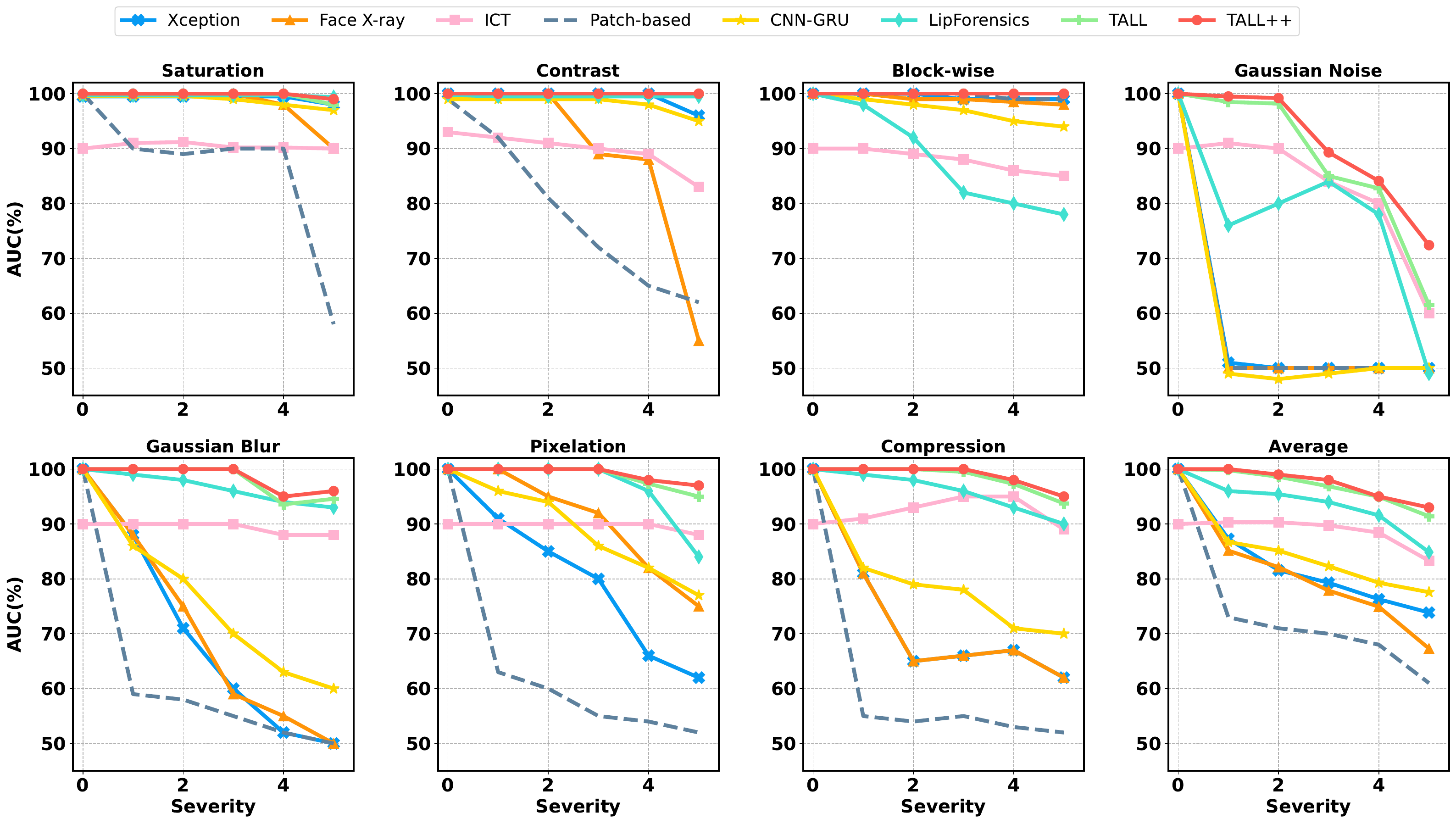}
\vspace{-1em}
\caption{\textbf{Robustness to unseen perturbations.} 
The Video-level AUC($\%$) fold visualization on DFo involves testing across five levels of seven corruptions. 1 to 5 represent a gradual increase in perturbation. “Average'' denotes the mean across all corruptions at each severity level.
}
\label{fig:rob}
\end{figure*}

\begin{figure*}[htbp]
\centering
\includegraphics[width=0.98\textwidth]{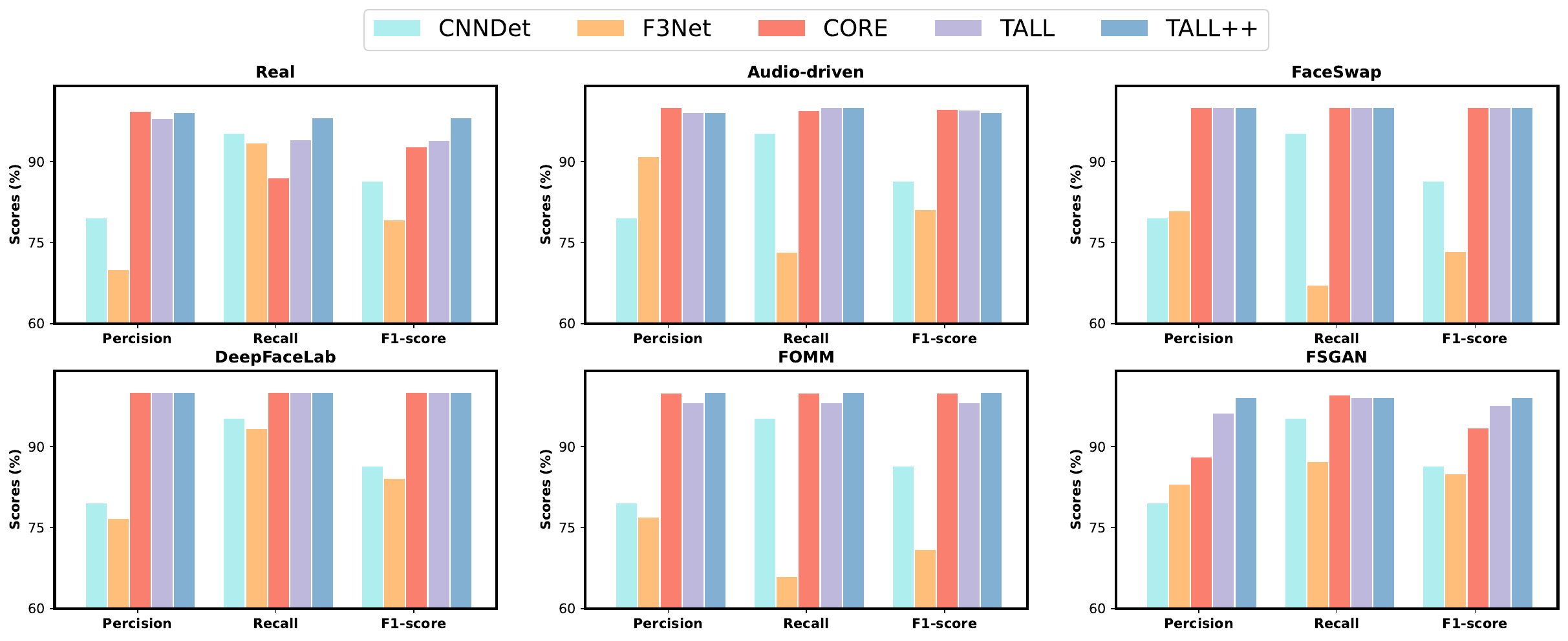}
\caption{Multi-classification Precision/Recall/F1 Scores($\%$) on KoDF dataset.
}
\label{fig:kodf}

\end{figure*}

\begin{table*}[ht]
\centering
\caption{\textbf{Average robustness to unseen perturbations.} The Video-level AUC($\%$) averaged across all severity levels for seven perturbations: Saturation, Contrast, Block-wise distortions, white Gaussian noise, Gaussian blurring, Pixelation, and Video Compression.
“Avg'' indicates the mean across all corruptions and all severity levels.}
\begin{tabular}{l c c c c c c c c c c c}
\toprule
Method  & {Clean} & Saturation & Contrast & Block & Noise & Blur & Pixel & Compress & Avg.  \\ 
\toprule
Xception~\cite{xception} & {99.8} & 99.3 & 98.6 & 99.7 & 53.8 & 60.2 & 74.2 & 62.1 & 78.3 \\
CNN-GRU~\cite{sabir2019gru} & {99.9} & 99.0 & 98.8 & 97.9 & 47.9 & 71.5 & 86.5 & 74.5 & 82.3 \\
CNN-aug~\cite{wang2020cnnaug}  & {99.8} & 99.3 & 99.1 & 95.2 & 54.7 & 76.5 & 91.2 & 72.5 & 84.1  \\
Patch-based~\cite{chai2020patch} & {99.9} & 84.3 & 74.2 & 99.2 & 50.0 & 54.4 & 56.7 & 53.4 & 67.5   \\
Face X-ray~\cite{FaceXray}  & {99.8} & 97.6 & 88.5 & 99.1 & 49.8 & 63.8 & 88.6 & 55.2 & 77.5  \\
LipForensics~\cite{haliassos2021lips} & 99.9 & 99.9 & 99.6 & 87.4 & 73.8 & 96.1 & 95.6 & 95.6 & 92.5  \\
FTCN~\cite{zheng:2021}  & 99.4 & 99.4 & 96.7 & 97.1 & 53.1 & 95.8 & 98.2 & 86.4 & 89.5 \\
RealForensics~\cite{haliassos2022realforensics} & 99.8 & 99.8 & 99.6 & 98.9 & 79.7 & 95.3 & 98.4 & 97.6 & 95.6 \\
\textbf{TALL} &  \textbf{100.0} & \textbf{100.0} & \textbf{100.0} & \textbf{100.0} &85.3 & 97.6 & 98.5 & 98.1 & 97.1  \\
\textbf{TALL++} &  \textbf{100.0} & \textbf{100.0} & \textbf{100.0} & \textbf{100.0} & \textbf{88.7}& \textbf{98.2} & \textbf{98.7}  & \textbf{98.6} &\textbf{98.0}  \\
\bottomrule
\end{tabular}
\label{tab:robustness}
\end{table*}

\subsection{Robustness Evaluation} 
In addition to demonstrating strong performance within intra-dataset and cross-dataset tasks, it is imperative that detectors exhibit robustness against common perturbations commonly encountered in social media videos.
To assess the robustness of our proposed methods, we adopt a testing methodology inspired by LipsForensics~\cite{haliassos2021lips}, exposing our models to testing across five severity levels and seven distinct perturbation types.
In Fig.~\ref{fig:rob}, we present a visual representation of how the increasing severity of each perturbation affects the performance of various methods. As anticipated, all methods show a decline in performance as perturbations become more severe. 
In comparison to the baseline method, TALL consistently delivers competitive performance across various perturbations. Notably, even when subjected to the most severe perturbation level of five, TALL maintains its superiority across all seven perturbation types.
When contrasted with TALL, TALL++ exhibits noticeable performance enhancements in Gaussian Noise, where it demonstrates significant improvements. Besides, we give the average AUC scores across all severities for each perturbation, as shown in Table~\ref{tab:robustness}, which further underscores our methods' remarkable robustness to a wide array of perturbations compared to previous methods.

\subsection{Deepfake Method Recognition}
Due to the constant emergence of new deepfake technologies and sophisticated videos each year, it is especially important for detectors to be able to recognize different deepfake generation methods. 
For the evaluation on KoDF, we sampled 7200, 1400, and 1400 videos for training, validation, and testing, while ensuring that individuals who appeared in the training set did not appear in the validation and testing sets. In the training set, there are 1200 videos of each different forgery type.

We conduct a comparative analysis of our method against previous approaches, all of which are replicated from the official open repository. 
Several noteworthy findings from this comparison are visually presented in Fig.~\ref{fig:kodf}. First, it's evident that our methods exhibit minimal fluctuations in all 3 metrics across 6 categories, signifying the consistent and robust performance of our method across all categories, even when faced with challenging ones like Real and FSGAN. 
In contrast, the precision of the consistently performing CORE~\cite{Ni_2022_CVPR} experiences a noticeable decline within the Real and FSGAN categories. Second, both our method and CORE achieve 100$\%$ precision, recall, and F1 scores in the FaceSwap and DeepFaceLab categories, indicating that these two types are the most readily identified. 
Third, TALL++ achieves the highest recall in the Real category, while genuine samples exhibit a higher susceptibility to being misclassified as fake. Overall, these findings underscore the reliability of our proposed methods.
Thirdly, genuine samples exhibit a higher susceptibility to being misclassified as fake, while TALL++ achieves the highest recall in the Real category. This is in contrast to CNNDet~\cite{wang2022adt}, CORE, and TALL, all of which demonstrate a decrease in recall. Overall, these findings underscore the reliability of our proposed methods.

\begin{figure}[htbp]
\centering
\includegraphics[width=0.48\textwidth]{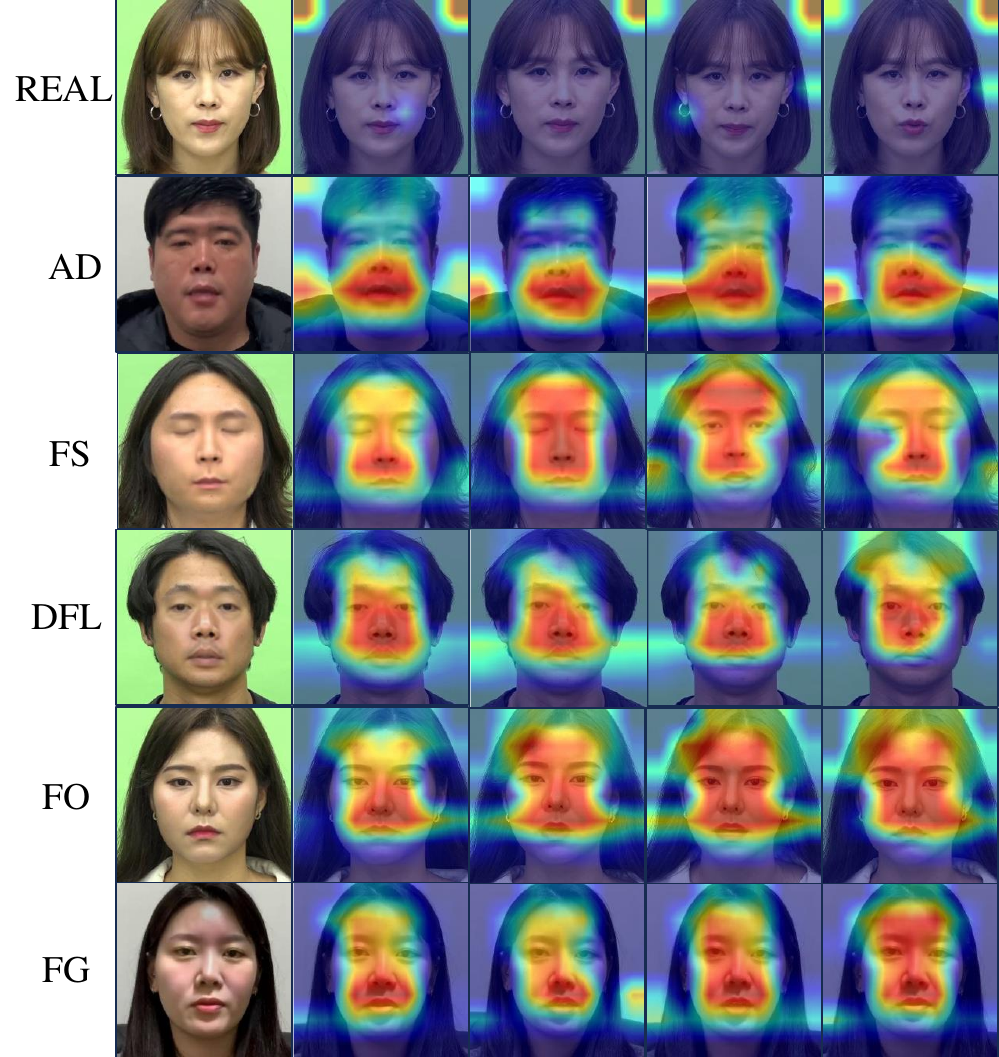}
\caption{Saliency map visualization on KoDF. {We give the class activation maps of a clip from the real videos and five types of fake videos (AD, FS, DFL, FO, FG) in the KoDF dataset. It can be observed that TALL++ can locate the forged areas corresponding to different forgery categories. For example, AD only manipulates the mouth region, FS swaps the square area of the source face to the corresponding area of the target face. }}
\label{cam:kodf}
\vspace{-1em}
\end{figure}

\begin{table*}[ht]
\centering
\small
\caption{Comprehensive comparisons of our methods and other deepfake detection methods on the DLB dataset. TALL$++^*$ indicates TALL without SC loss.}
\renewcommand{\arraystretch}{1.2}
\resizebox{1.\linewidth}{!}{
\begin{tabular}{lcccccccccccc}
\toprule
\multirow{2}{*}{Method} & Training & \multicolumn{10}{c}{Testing dataset} & \multicolumn{1}{c}{Total} \\ \cline{3-12}
 &
  dataset &
  \multicolumn{1}{c}{PGAN} &
  \multicolumn{1}{c}{SGAN} &
  \multicolumn{1}{c}{PjGAN} &
  \multicolumn{1}{c}{D-SGAN2} &
  \multicolumn{1}{c}{D-PjGAN} &
  \multicolumn{1}{c}{DDPM} &
  \multicolumn{1}{c}{IDDPM} &
  \multicolumn{1}{c}{ADM} &
  \multicolumn{1}{c}{PNDM} &
  \multicolumn{1}{c}{LDM} &
  \multicolumn{1}{c}{Avg.} \\ \toprule
CNNDet~\cite{wang2020cnnaug}      & FF++     & 50.93  & 51.33  &  49.55 & 45.96  & 50.47  & 49.26  &  50.64 & 50.52  & 46.58  & 53.15 & 49.84                           \\
F3Net~\cite{qian2020}                    & FF++     &  50.25 &  49.02  & 48.03 &56.81   & 50.05 & 49.51  & 53.14  &  49.25 & 56.37  & 52.29 & 51.47                         \\
CORE~\cite{Ni_2022_CVPR}                    & FF++     & 54.16  & 47.72  &  47.85 &  59.60   & 46.61  & 58.23  &  58.39   & 54.18   & 62.38  & 58.15  & 54.73                           \\
\textbf{TALL}                & FF++     & 67.17  & 58.51  & 65.30  & 70.14  & 67.18  & 71.31  & 73.12  & 69.65  & 75.71  & 65.24 & 68.33                          \\ 
\textbf{TALL${++}^*$}                & FF++     & 70.90  &  62.33 & 69.76  & 73.98 & 69.30  & 76.41  & 78.36  & 76.43  & 78.62  & 68.98 & \textbf{72.51}                           \\ \midrule
CNNDet~\cite{wang2020cnnaug}                  & GANs     & 99.99  & 99.99  & 100.00 &100.00   & 99.99  & 70.47  & 80.18 & 81.72  & 99.99  & 93.66 & 92.60                           \\
F3Net~\cite{qian2020}                    & GANs     &97.92   & 98.64  & 90.36  & 98.12  & 87.26  & 89.11  & 86.86  & 73.58  & 98.55  & 95.49 & 91.59                          \\
CORE~\cite{Ni_2022_CVPR}                       & GANs     & 99.24  & 99.47  & 99.34  & 99.93  & 99.93  & 83.66  & 82.18  &  76.96 & 99.51  & 98.77 & 93.90                          \\
TALL                & GANs    & 100.00   &100.00    & 100.00   & 100.00   &100.00    &95.76    & 96.88   & 89.75   & 100.00   & 100.00  &   98.23                         \\ 
\textbf{TALL${++}^*$}               & GANs     & 100.00  & 100.00  & 100.00  & 100.00  & 100.00   &100.00    & 100.00   & 97.53   & 100.00   & 100.00  & \textbf{99.75}                           \\ \midrule
CNNDet~\cite{wang2020cnnaug}                  & DMs      &  99.98 & 99.97  &99.98   &  99.94 &  99.99 &99.99   &  99.99 & 99.98  & 99.99   & 99.99 &  95.36                         \\
F3Net~\cite{qian2020}                   & DMs      &96.56   & 96.47  & 91.74  & 90.54  & 98.79   &95.17   & 94.97  & 90.08  & 99.86  & 99.46 &   99.78                        \\
CORE~\cite{Ni_2022_CVPR}                       & DMs      & 99.77 &  99.88 & 99.56  & 99.96  & 99.63  &99.95   & 99.85  & 99.44  & 99.99  & 99.97 & 99.78                          \\
TALL                & DMs    &100.00   &100.00   &99.99   & 99.97  &100.00   & 100.00  & 100.00  & 100.00  &100.00   &100.00  &99.99                         \\ 
\textbf{TALL${++}^*$}              & DMs      &100.00   &100.00   & 100.00  & 100.00  &  100.00 &100.00   &100.00  &  100.00 & 100.00  & 100.00 &  \textbf{100.00}                         \\ \bottomrule

\end{tabular}
}
\label{table:dms}
\end{table*}

\textbf{Saliency map visualization.}
Fig.~\ref{cam:kodf} visualizes the saliency map of each category on KoDF. We can observe that the model pays more attention to the background regions of the real images and also demonstrates the discriminative capability for different categories. 
For instance, it can respond to the distinct boundaries in the motion-driven Fo category and the AD category at their respective positions with relatively clear delineations.

\subsection{Detection of DM \& GAN-Generated Images}
Recently, Diffusion Models (DMs) have established a novel image generation paradigm owing to their impressive capacity for producing high-quality images. 
Although previous forgery detectors have shown promising performance, their generalization ability is unsatisfactory when confronted with images generated by recent DMs and full-image generation using GANs.
Following DDMD~\cite{ricker2022towards}, we delve into the generalizability of our approach using the DBL dataset, composed of images entirely generated by GANs and DMs. 
{Note that DBL dataset only contains image data. 
To evaluate our method on this dataset, we can only conduct experiments at the image level. 
Under the image-level scheme, the semantic consistency loss we proposed is not applicable, thus we set its parameter$\alpha$ to zero. 
However, in order to keep the same experimental settings and give video-level results, we assume a video is composed of 200 images.
Then we average the model predictions across every 200 images.}
As shown in Table~\ref{table:dms}, we include three scenarios, FF++$\rightarrow$GANs$\&$DMs, GANs$\rightarrow$GANs$\&$DMs and DMs$\rightarrow$GANs$\&$DMs. Besides, we reproduce previous detectors including CNNDet, F3Net, and CORE with the official codes to compare with our methods. 

We first show the results of models trained on the FF++ (HQ) datasets and tested on 10 generated categories images of the DBL dataset. It is expected that previous methods experience a significant drop in performance when dealing with images generated through diffusion or GAN full-maps, resulting in an average AUC lower than 55$\%$.
In contrast, TALL++ works well, advancing the average accuracy from 68.33$\%$ in TALL to 72.51$\%$.
We also include GANs-generated images and DMs-generated images as training data and
verify the models under GANs$\rightarrow$ GANs$\&$DMs and DMs$\rightarrow$GANs$\&$DMs scenarios.
In Table~\ref{table:dms}, our methods obtain the best AUC for each category. 
Notably, when compared with TALL, TALL++ consistently outperforms it, which can likely be attributed to the contribution of GRB.
Besides, we can easily find that CORE~\cite{Ni_2022_CVPR} trained on DM-generated data attains an impressive 99$\%$ when evaluated on GAN-generated data. In contrast, when GAN-generated images are employed as training data, the AUC values on IDDPM and ADM are notably lower, standing at 82.18$\%$ and 76.96$\%$, respectively.

\subsection{Model Analysis and Discussions}
\textbf{Scaling over different backbones.}
To assess the efficacy of TALL, we have employed a range of commonly utilized image-level backbones for deepfake detection and contrasted them with video-level backbones.
As shown in Table~\ref{tab:backcbone}, TALL scales well to image-based backbones and transformers, all without imposing additional computational burdens. 

We first compare the accuracy and complexity of the CNN-based video and image backbones. 
Although I3D~\cite{carreira2017i3d} and R3D~\cite{hara2017learning} achieve better performance than vanilla ResNet50~\cite{He_2016_CVPR} and EfficientNet~\cite{tan2019efficientnet}, the computation costs are huge, such as R3D-50 with 296G FLOPs.
TALL expansion to Resnet and EfficentNet results in a 3.6$\%$ AUC boost on the CDF for ResNet and a 1.5$\%$ AUC boost on the CDF for EfficentNet.

We further extend TALL to several transformers in the second section.
The table provides a clear illustration of TALL consistently enhancing accuracy across various backbones and transformers. Furthermore, TALL++ further elevates AUC from 76.78$\%$ to 77.51$\%$ on challenging DFDC dataset, indicating the effectiveness of GRB and SC Loss.

\begin{table}[ht]
\centering
\setlength\tabcolsep{2pt}
\caption{\textbf{Scaling over backbones with TALL.} TALL consistently improves the accuracy over different image-level models. We show the AUC, FLOPs, and number of parameters for each model on the cross-dataset scenario. All models are trained on FF++ (HQ). \checkmark~indicates the model enables temporal modeling. * indicates our implementation. PT indicates pre-train. 1K and 21K indicate the model pre-trained on ImageNet-1K and 21K respectively. } 
\begin{tabular}{lcccccc}
\toprule
Models   & Temp. & CDF   & DFDC  & FLOPs  & Params & PT \\ \midrule
I3D-RGB$^*$~\cite{carreira2017i3d}  & \checkmark     & 78.24 & 65.58 & 222.7G & 25M     & 1K     \\ 
R3D-50$^*$~\cite{hara2017learning}    & \checkmark    & 79.63 & 67.73 & 296.6G & 46M     & 1K     \\ \midrule
ResNet50$^*$~\cite{He_2016_CVPR} & $\times$      & 76.38 & 64.01 & 25.5G  & 21M     & 1K     \\ 
\textbf{+TALL}    & \checkmark     & 80.90   & 65.54    & 25.5G  & 21M     & 1K     \\ 
\textbf{+TALL++}    & \checkmark     & 81.43   & 67.10    & 35.5G  & 23M     & 1K     \\ \midrule
EffNetB4$^*$~\cite{tan2019efficientnet} & $\times$      & 78.19 & 66.81 & 8.3G   & 19M     & 1K     \\
\textbf{+TALL}    & \checkmark     & 83.37     & 67.15  & 8.3G   &  19M   & 1K     \\ 
\textbf{+TALL++}    & \checkmark     & \textbf{84.71}     & \textbf{68.37}  & 16.3G   &  21M   & 1K     \\\midrule \midrule
VTN~\cite{neimark2021video}      & \checkmark     & 83.20 & 73.50 & 296.6G &46M    & 21K    \\
VidTR~\cite{zhang2021vidtr}    & \checkmark     & 83.30 & 73.30 & 117G   & 93M     & 21K    \\
ViViT$^*$~\cite{vivit}    & \checkmark     & 86.96 & 74.61 & 628G   & 310M    & 21K    \\
ISTVT~\cite{zhao2023istvt}    & \checkmark     & 84.10 & 74.20 & 455.8G & -       & -        \\ \midrule
ViT-B$^*$~\cite{vit}    & $\times$      & 82.33 & 72.64 & 55.4G  & 84M     & 21K    \\
\textbf{+TALL}    & \checkmark     & 86.58    & 74.10     & 55.4G  &  84M    & 21K    \\
\textbf{+TALL++}    & \checkmark     & 87.14    & 75.93     & 60.4G  &  86M    & 21K    \\\midrule
Swin-B$^*$~\cite{liu2021swin}   & $\times$      & 83.13 & 73.01 & 47.5G  & 86M     & 21K    \\
\textbf{+TALL}   & \checkmark     & 90.79 & 76.78 &  47.5G  & 86M     & 21K    \\
\textbf{+TALL++}   & \checkmark     & \textbf{91.96} & \textbf{78.51} & 67.5G & 88M     & 21K    \\ \bottomrule
\end{tabular}
\label{tab:backcbone}
\end{table}

\textbf{Ablation study of model components.}
We ablate the different components of our method quantitatively on CDF and DFDC datasets in Table~\ref{exp:abl}. 
The baseline model in our ablation study is the Swin-B that predicts the video-level deepfake label by minimizing the following standard cross-entropy loss.
Several noteworthy results can be found in Table~\ref{exp:abl}. 
First, both TALL and its mask demonstrate improvements in AUC scores on both the CDF and DFDC datasets, underscoring the effectiveness of our proposed TALL strategy.
Second, we observe that GRB and SC Loss contribute significantly to cross-dataset tasks, highlighting the valuable role of high-level semantic information in enhancing model generalization.
Third, the GRB's ability to reason with high-level semantic information proves especially beneficial when dealing with the challenging DFDC dataset. In comparison to variant 3, variant 6 exhibits a 1$\%$ improvement on DFDC, while showing a 0.6$\%$ improvement on the CDF dataset.
Generally, each component of TALL++ is beneficial to deepfake detection.

\begin{table}[ht]
\centering
\caption{\textbf{Ablation study of each component of TALL++.} The baseline model is the Swin-B with cross-entropy loss. AUC ($\%$) of different variants for the cross-dataset task on CDF and DFDC datasets.} 
\setlength\tabcolsep{3.5pt}
\begin{tabular}{c|ccccc|cc}
\toprule
No. & Baseline & TALL & Mask & GRB & SC Loss               & CDF & DFDC \\ \midrule
1 &\Checkmark     & \XSolidBrush   & \XSolidBrush    & \XSolidBrush    & \XSolidBrush & 83.13    & 73.01     \\
2 &   \Checkmark    &  \Checkmark    &   \XSolidBrush   &   \XSolidBrush  & \XSolidBrush & 87.13    & 74.32     \\
3 &  \Checkmark      &  \Checkmark     &  \Checkmark     & \XSolidBrush    & \XSolidBrush & 90.79    &   76.78   \\
4   & \Checkmark     & \XSolidBrush      & \XSolidBrush       & \Checkmark       & \XSolidBrush  &  83.61    & 74.13     \\ 
5   & \Checkmark     & \XSolidBrush      & \XSolidBrush       & \Checkmark       & \Checkmark  & 83.99    & 74.38     \\
6 &  \Checkmark     & \Checkmark       & \Checkmark       & \Checkmark       & \XSolidBrush & 91.42    & 77.80     \\  
7 &   \Checkmark     &\Checkmark      & \Checkmark       & \Checkmark       & \Checkmark  &  \textbf{91.96}  & \textbf{78.51}     \\ 
  \bottomrule
\end{tabular}
\label{exp:abl}
\end{table}

\subsection{Design analysis of TALL}
\begin{figure}[htbp]
    \centering
     \captionof{table}{AUC ($\%$) of different layouts on CDF and DFDC for the cross-dataset task.}
	\begin{minipage}{0.45\linewidth}
		\centering
		\includegraphics[width=\linewidth]{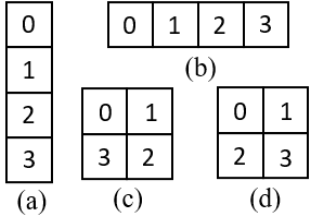}

	\end{minipage}

	\begin{minipage}{0.45\linewidth}
	    \centering
            \small
            \setlength\tabcolsep{3.5pt}
            \begin{tabular}{cll}
            \toprule
            Layout & CDF & DFDC \\ \midrule
             (a)    & 88.36    & 75.48   \\
             (b)    & 88.50     & 75.63   \\
             (c)    & 89.11    & 77.51  \\
             (d)    &\textbf{91.96}  & \textbf{78.51}   \\ \bottomrule  
            \end{tabular}
        \label{fig:layout}
	    \end{minipage}

\end{figure}

\textbf{Discussion on different layouts.} 
In this section, we perform cross-dataset experiments utilizing various layouts and analyze which layout fosters the most robust model generalization. The results, as presented in Table~\ref{fig:layout}, demonstrate that the more densely arranged layouts (c) and (d) outperform the single-dimensional arrangements (a) and (b). We attribute this performance improvement to the closer proximity of every two frames in the compact layout, which aids the model in capturing local differences between frames. Furthermore, we find that frames arranged as shown in layout (d) achieved the best cross-dataset AUC scores.

\textbf{Discussion on different orders.}
In this case, we initially explore the impact of omitting the last sub-image and the last two sub-images on the model's performance.
The first two rows of the table show that multi-frame aggregation enhances the performance of cross-dataset tasks.
Furthermore, we examine different orders of TALL's sub-frames to evaluate TALL++, considering three arrangements: forward, reverse, and random.
Results are shown in Table~\ref{table:order}, the forward order performs the best for three different orders. 
This superiority might be attributed to the temporal position encoding within TALL++.

\begin{table}[ht]
\centering
\caption{AUC ($\%$) of different orders on CDF and DFDC for cross-dataset task.}
\setlength\tabcolsep{10pt}{
\begin{tabular}{ccc}
\toprule
Variants  & CDF & DFDC \\ \midrule
0, 1, 2, -    &   89.73    & 74.46   \\ 
0, 1, -, -    &   88.09     &  73.95   \\ 
Random      &   88.34     & 76.23    \\
Reverse    &   90.12   & 77.75 \\
Forward    &   \textbf{91.96}      & \textbf{78.51} \\
\bottomrule  
\end{tabular}
}
\label{table:order}
\end{table}

\textbf{Discussion on the sub-frame size.}
To better understand the effects of sub-frame size and explore the possibility of containing more video frames, we test several variants in the cross-dataset task on CDF and DFDC datasets, shown in Table~\ref{tb:size}.
We observe that when we transition the layout to a 3x3 grid and increase the sub-frame size to 224x224, there is a modest increase of +1.17$\%$ in the AUC on the CDF dataset when compared to our default setting (the fifth row) of 91.96$\%$. However, the computational costs, measured in FLOPs, increase almost fivefold to 47.5G. 
Subsequently, we proceed to evaluate alternative configurations, with the aim of retaining the thumbnail size at 224x224 while reducing the dimensions of the sub-frame. Our analyses clearly demonstrate that such adjustments lead to a discernible deterioration in performance across both the CDF and DFDC datasets. 
Specifically, the layout configuration of 2x2 with a sub-frame size of 112x112 emerges as the optimal choice, striking an optimal balance between performance and computational costs.

\begin{table}[ht]
\centering
\caption{{AUC ($\%$) of the variants of thumbnail's size on CDF and DFDC for cross-dataset task. DS indicates downsampling, and size indicates the size of a thumbnail. }}
\begin{tabular}{ccccccc}
\toprule
DS & Size & Layout & FLOPs  & CDF & DFDC            \\ \midrule
False  &[672,672] & ${3}\times{3}$ & 540G  & 93.13 & 79.84  \\
False  &[448,448] & ${2}\times{2}$ & 290G  & 92.10  & 78.63 \\
3$\times$   &[224,224] & ${3}\times{3}$ & 67.5G &85.57 &75.51 \\
4$\times$    &[224,224] & ${4}\times{4}$ & 67.5G &82.31 &73.46 \\
2$\times$  &[224,224] & ${2}\times{2}$ & 67.5G  &91.96  &78.51 \\ \bottomrule  
\end{tabular}
\label{tb:size}
\end{table}

\textbf{Discussion on the window size.} 
We investigate the AUC of different shifted window sizes within the training process. The results are detailed in Table~\ref{table:window}. Expanding the window size from Swin-B's default value of 7 to 14 during the initial three phases leads to a significant enhancement in model performance. This augmentation improves 3.72$\%$ and 2.5$\%$ in AUC on the CDF and DFDC datasets, respectively. However, further increasing the window size to 28 does not improve performance as shown in the third row. We analyze that an excessively large window may impede the model's ability to capture local information within the sub-frame.

\begin{table}[ht]
\centering
\caption{AUC ($\%$) of the different window sizes of Swin-B on CDF and DFDC for cross-dataset task.}
\setlength\tabcolsep{10pt}
\begin{tabular}{ccc}
\toprule
Window size   & CDF & DFDC  \\ \midrule
(7,7,7,7)     & 88.24   &  76.01     \\
(28,28,28,7) & 89.26    &   76.43     \\
(14,14,14,7)    &\textbf{91.96}  &  \textbf{78.51}    \\
 \bottomrule
\end{tabular}
\label{table:window}
\vspace{-3em}
\end{table}

\section{Conclusion}

This paper presents a novel approach, termed TALL, for face deepfake video detection. {TALL converts video clips into thumbnails that capture both spatial and temporal information}, facilitating the effective identification of spatiotemporal inconsistencies.
We further introduce a graph reasoning block and semantic consistency loss to enhance TALL, resulting in TALL++, which leverages high-level semantic features to improve generalization capabilities. TALL++ achieves the best trade-off between deepfake detection accuracy and computational costs.
Experiments on intra-dataset, cross-dataset, diffusion-generated image evaluations, and deepfake generation method recognition verify that TALL and TALL++ achieve results comparable to or even better than the state-of-the-art methods.

\section*{Data Availability Statement.}
All the datasets used in this paper are available online. 
FaceForensics++ (FF++)~\footnote{\url{https://github.com/ondyari/FaceForensics}}, Celeb-DF (CDF)~\footnote{\url{https://github.com/yuezunli/celeb-deepfakeforensics}}, 
DFDC~\footnote{\url{https://ai.meta.com/datasets/dfdc/}}, DeeperForensics (DFo)~\footnote{\url{https://liming-jiang.com/projects/DrF1/DrF1.html}}, 
FaceShifter(Fsh)~\footnote{\url{https://github.com/ondyari/FaceForensics}},
Wild-Deepfake (Wild-DF)~\footnote{\url{https://github.com/deepfakeinthewild/deepfake-in-the-wild}}, KoDF~\footnote{\url{https://deepbrainai-research.github.io/kodf/}}, and Deepfakes LSUN-Bedroom (DBL)~\footnote{\url{https://github.com/jonasricker/diffusion-model-deepfake-detection\#dataset}} can be downloaded from their official website accordingly. 

\section*{Acknowledgment}
The authors would like to thank the reviewers and the associate editor for their valuable comments.
The authors also thank Ziming Yang, and Gengyun Jia for their help in improving the technical writing aspect of this paper.

{
\bibliographystyle{unsrt}
\bibliography{reference}}

\end{document}